\documentclass{article}
\usepackage{graphicx}
\usepackage{float}
\usepackage{booktabs}
\usepackage{tikz}
\usepackage{comment}
\usepackage{amsmath,amssymb}
\usepackage{color}
\usepackage{subcaption}
\usepackage{breakcites}
\usepackage{pifont}

\everypar{\looseness=-1}

\definecolor{ForestGreen}{RGB}{34,139,34}

\usepackage{xcolor, soul}
\usepackage{xfrac}
\usepackage{multirow}

\usepackage[inline]{enumitem}
\usepackage[colorlinks]{hyperref}
\usepackage[capitalize]{cleveref}
\usepackage[preprint,nonatbib]{preprint_2022}

\usepackage{enumitem}
\newlist{properties}{enumerate}{10}
\setlist[properties]{label*=\arabic*}
\crefname{propertiesi}{property}{Properties}
\Crefname{propertiesi}{property}{Properties}

\title{ContraCLIP: Interpretable GAN generation driven by pairs of contrasting sentences}

\usepackage[utf8]{inputenc} 
\usepackage[T1]{fontenc}    
\usepackage{hyperref}       
\usepackage{url}            
\usepackage{booktabs}       
\usepackage{amsfonts}       
\usepackage{nicefrac}       
\usepackage{microtype}      
\usepackage{xcolor}         

\usepackage[toc,page]{appendix}

\newcommand{\Affiliation}{%
\end{tabular}\\\begin{tabular}[t]{c}\ignorespaces%
}
\author{%
  Christos Tzelepis$^1$
  \And 
  James Oldfield$^1$
  \And
  Georgios Tzimiropoulos$^1$
  \AND
  Ioannis Patras$^1$
  \Affiliation\\
  $^1$School of Electronic Engineering and Computer Science, Queen Mary University of London
}

\begin{document}

\maketitle

\begin{abstract}
\looseness-1This work addresses the problem of discovering non-linear interpretable paths in the latent space of pre-trained GANs in a model-agnostic manner. In the proposed method, the discovery is driven by a set of pairs of natural language sentences with contrasting semantics, named \textit{semantic dipoles}, that serve as the ``limits'' of the interpretation that we require by the trainable latent paths to encode. By using the pre-trained CLIP encoder~\cite{clip-radford2021icml}, the sentences are projected into the vision-language space, where they serve as dipoles, and where RBF-based warping functions define a set of non-linear directional paths, one for each semantic dipole, allowing in this way traversals from one semantic pole to the other. By defining an objective that discovers paths in the latent space of GANs that generate changes along the desired paths in the vision-language embedding space, we provide an intuitive way of controlling the underlying generative factors and address some of the limitations of the state-of-the-art works, namely, that a) they are typically tailored to specific GAN architectures (i.e., StyleGAN), b) they disregard the relative position of the manipulated and the original image in the image embedding and the relative position of the image and the text embeddings, and c) they lead to abrupt image manipulations and quickly arrive at regions of low density and, thus, low image quality, providing limited control of the generative factors. We provide extensive qualitative and quantitative results that demonstrate our claims with two pre-trained GANs, and make the code and the pre-trained models publicly available at: \url{https://github.com/chi0tzp/ContraCLIP}.
\begin{figure}[h]
\centering
\includegraphics[width=\linewidth]{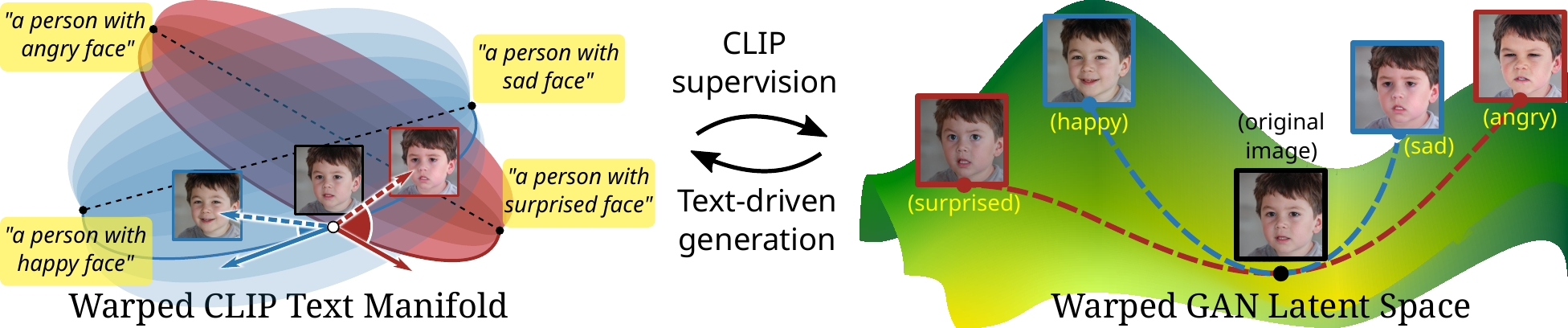}
\caption{The CLIP~\cite{clip-radford2021icml} text space, warped due to \textit{semantic dipoles} of contrasting pairs of sentences in natural language, provides supervision to the optimisation of non-linear interpretable paths in the latent space of a pre-trained GAN.}
\label{fig:summary}
\end{figure}
\end{abstract}

\section{Introduction}\label{sec:intro}
    
    During the recent years, Generative Adversarial Networks (GANs)~\cite{goodfellow2014generative} have emerged as the dominating generative learning paradigm for the task of image synthesis, and they continue to dramatically change a wide and diverse range of disciplines, such as super-resolution~\cite{ledig2017photo}, face restoration~\cite{yang2021gan}, and editing~\cite{talktoedit2021iccv}, while they have also been incorporated in discriminative tasks~\cite{xu2021generative}. Despite their remarkable capability of capturing and modeling image data distributions through a semantically rich latent space, traversing a pre-trained GAN's latent space in an interpretable/controllable manner remains an open problem.
    
    Exploring the latent space of a pre-trained GAN in an interpretable manner has drawn significant attention from the research community during the recent years~\cite{ganspace2020harkonen,voynov2020unsupervised,sefa2021cvpr,oldfield2021tensor,warpedganspace2021iccv,goetschalckx2019ganalyze,shen2020interpreting,plumerault20iclr,jahanian20iclr,abdal2021styleflow,wu2021stylespace,shen2020interfacegan}. Typically, these works first discover a set of linear or non-linear latent paths, in an unsupervised or (semi)supervised manner, and then try to label them based either on laborious manual annotation (e.g.,~\cite{voynov2020unsupervised,ganspace2020harkonen}) or by incorporating pre-trained detectors (e.g., smile detector)~\cite{warpedganspace2021iccv}. While using pre-trained detectors may label effectively the discovered latent paths, this process might ignore certain generative factors present in the GAN latent space since a) there is limited availability of pre-trained detectors for attributes of interest for the given pre-trained GAN, and b) pre-trained detectors are inherently limited to a closed set of visual concepts/attributes. 
    
\begin{figure}[t]
        \centering
        \includegraphics[width=\textwidth]{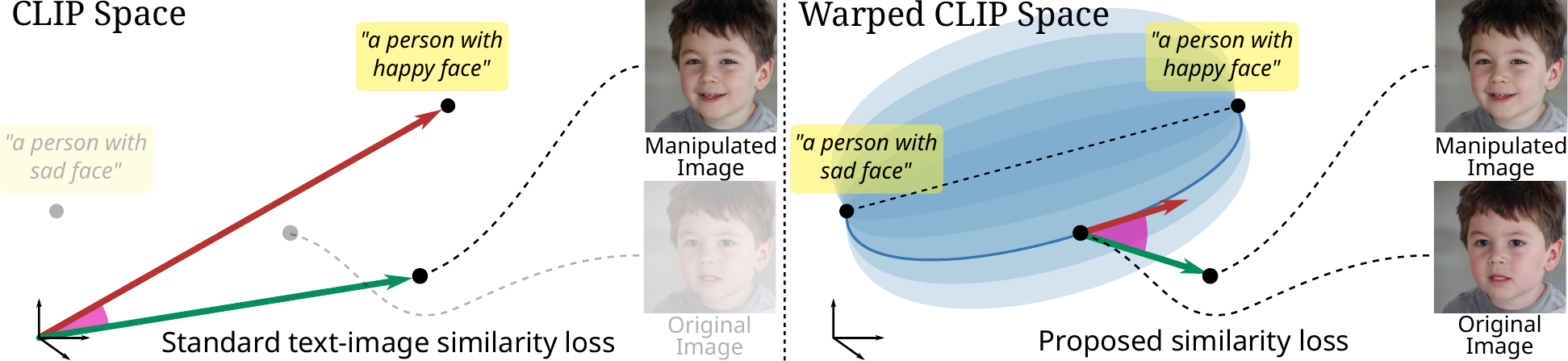}
        \caption{Methodological comparison between the proposed and the standard text-image cosine similarity loss calculation. Standard text-image similarity considers the features of a given image and the features of a given text prompt (encoded by CLIP) and calculates their cosine distance. We propose the use of local text directions along the non-linear paths that are induced by the warping of the text space due to a pair of contrasting sentences (semantic dipole).}
        \label{fig:standard_vs_proposed_sim}
    \end{figure}

    To address the above limitations, a few recent works have incorporated joint vision-language models due to their inherent capability of expressing a much wider set of visual concepts. TediGAN~\cite{xia2021tedigan} proposes a novel GAN inversion technique that can map multi-modal information (i.e., texts, sketches, labels) into a common latent space of a pre-trained StyleGAN~\cite{stylegan2_karras20cvpr}. Then, using visual-linguistic similarity learning, it provides an interactive system where the user provides input textual guidelines and the system generates diverse images given the same input text, allowing the user to edit the appearance of different attributes interactively. However, for doing so, it relies on both annotated training data and the hierarchical structure of the StyleGAN architecture. More specifically, the authors of~\cite{xia2021tedigan} introduce and use the Multi-Modal CelebA-HQ, a curated dataset of face images, that is augmented with a high-quality segmentation mask, sketch, and descriptive text. Then, they rely on the hierarchical characteristic of StyleGAN's $\mathcal{W}$-space (i.e., style-mixing) to learn the text-image matching by mapping the image and text into the same joint embedding space. By contrast, our method is not limited to the task of face editing, nor tied to a specific GAN architecture (i.e., StyleGAN) and, thus, can be applied to any latent space. Another recent work, closely related to ours, is StyleCLIP~\cite{patashnik2021styleclip}, which utilizes a CLIP~\cite{clip-radford2021icml}-based similarity loss in order to learn a StyleGAN-specific latent mapper network that infers a text-guided latent manipulation step for a given input image. To do so, StyleCLIP uses the CLIP text features of the given text prompt as the target vector and tries to align it with the CLIP image features of the manipulated image, adopting the standard text-image cosine similarity method (left part of Fig.~\ref{fig:standard_vs_proposed_sim}). However, the adopted loss criterion disregards a) the relative position of the manipulated and the original image in the image embedding, and b) the relative position of the image and the text embeddings since all are expressed in relation to the origin of the axes in the CLIP embedding space. This leads to abrupt image manipulations and quickly arrives at regions of low density and, thus, low image quality, as shown by the experimental results.
    
    In this work, we propose to learn non-linear interpretable paths in the latent space of a pre-trained GAN driven by \textit{semantic dipoles} given in natural language. More specifically, we define pairs of sentences, each pair corresponding to one trainable non-linear path in the GAN latent space and expressing contrasting meanings in natural language. When represented in the CLIP text space, the aforementioned pairs form ``dipoles'' that serve as the ``limits'' of the interpretation that we require by the optimised latent paths to encode. Then, we use the poles (i.e., the feature representation of each sentence in the pair) as the centres of an RBF-based warping function that endows the text space with a family of non-linear curves, providing end-to-end directional paths from the one pole/sentence to the other (Fig.~\ref{fig:summary}). By doing so, we decouple the supervisory text features from the origin of axes, establishing at the same time only local relations between the text and the image features. This leads to non-linear text paths that are smooth and allow for mild transition between the given point and the desired end. This is illustrated in the right part of Fig.~\ref{fig:standard_vs_proposed_sim}. As a result, transition to images described by the desired text prompts is smoother and longer than those of StyleCLIP~\cite{patashnik2021styleclip}, while traveling to regions of low density is significantly prevented, as will be shown in the experiments. The main contributions of this paper can be summarized as follows:
    \begin{itemize}
        \item We propose a method for traversing the latent space of a pre-trained GAN in an interpretable manner driven by semantic dipoles in natural language.
        \item We model non-linear paths in the CLIP text space which allow for smooth transition from one pole to the other, preventing low-density regions/artifacts.
        \item Our method is model-agnostic and not tied to any specific GAN architecture (e.g., StyleGAN), such as~\cite{xia2021tedigan} and~\cite{patashnik2021styleclip}.
        \item We apply our method to two different GAN architectures (i.e., ProgGAN~\cite{proggan_karras18iclr}, and StyleGAN2~\cite{stylegan2_karras20cvpr}) and compare with GANSpace~\cite{ganspace2020harkonen}, WGS~\cite{warpedganspace2021iccv}, and StyleCLIP~\cite{patashnik2021styleclip}. We show that in comparison to state-of-the-art vision-language StyleCLIP~\cite{patashnik2021styleclip}, our method produces longer and more disentangled interpretable paths, generating images of higher quality with far better control of the generative factors.
    \end{itemize}

\section{Related Work}\label{sec:related_work}
    
    \subsection{Interpretable latent paths in pre-trained GAN generators}
        Exploring the latent space of a pre-trained GAN in an interpretable manner has typically been approached by the research community by discovering a set of linear~\cite{voynov2020unsupervised,ganspace2020harkonen} or non-linear~\cite{warpedganspace2021iccv} latent paths, in an unsupervised or (semi)supervised manner, and then trying to label them based either on laborious manual annotation (e.g.,~\cite{voynov2020unsupervised,ganspace2020harkonen}) or by incorporating pre-trained detectors~\cite{warpedganspace2021iccv}. GANSpace~\cite{ganspace2020harkonen} performs PCA on deep features at the early layers of the generator and finds directions in the latent space that best map to those deep PCA vectors, arriving at a set of non-orthogonal directions in the latent space. Voynov and Babenko~\cite{voynov2020unsupervised} proposed an unsupervised method to discover linear interpretable latent space directions by requiring that the aforementioned directions lead to image transformations easily distinguishable from each other by a discriminator/reconstructor network. WarpedGANSpace (WGS)~\cite{warpedganspace2021iccv} extended this to the non-linear case by optimising a set of RBF-based functions, each giving rise to a family of interpretable paths in the latent space. However, whilst our method builds on them, the discovery of the paths in the latent space of GANs is instead driven by textual descriptions provided in natural language that define dipoles in the the CLIP text space.
    
    \subsection{Vision-language models}
        Cross-modal vision-language representations have recently drawn attention towards a plethora of related tasks, such as image captioning, visual question answering, and language-based image retrieval~\cite{li2020oscar,chen2020uniter,li2020unicoder,lu2019vilbert,sariyildiz2020learning,desai2021virtex}. This line of research has been revolutionized by the use of Transformers~\cite{vaswani2017attention,devlin2018bert}. DALL-E~\cite{ramesh2021zero}, a 12-billion parameter version of GPT-3~\cite{brown2020language}, has exhibited remarkable capabilities in generating and applying transformations to images guided by text, however, even in 16-bit precision, it requires over 24 GB of GPU VRAM. In~\cite{clip-radford2021icml}, Radford et al. introduced Contrastive Language-Image Pre-training (CLIP)~\cite{clip-radford2021icml}, a vision-language model pre-trained on 400 million image-text pairs collected from a variety of publicly available sources on the Internet. CLIP provides a vision-language embedding space that allows the estimation of the semantic similarity between given image-text pairs. Its powerful representation scheme (a Transformer~\cite{vaswani2017attention} as a Text Encoder and a Vision Transformer~\cite{dosovitskiy2020image} as an Image Encoder) exhibit state-of-the-art zero-shot image recognition performance. In this work, we use CLIP's joint text-image space, which we warp using RBF-based functions, in order to arrive at non-linear text paths that we use as a supervisory signal in order to optimize non-linear traversal paths in the GAN latent space.
    
    \subsection{Text-guided image generation and manipulation}
        Tracing back to~\cite{reed2016generative}, text-guided image generation has gained considerable attention. AttnGAN~\cite{xu2018attngan} incorporated an attention mechanism between text and image features in order to synthesize fine-grained details at different sub-regions of the image by paying attention to the relevant words in the natural language description. ManiGAN~\cite{li2020manigan} goes beyond AttnGAN by semantically editing parts of an image matching a given text that describes desired attributes (e.g., texture, colour, and background), while preserving other contents that are irrelevant to the text. In~\cite{talktoedit2021iccv} the authors propose Talk-to-Edit that performs interactive fine-grained attribute manipulation through dialog between the user and the system. To do so, they gather and curate a visual-language dataset of facial images, annotated with rich fine-grained labels, which classify one attribute into multiple degrees according to its semantic meaning, along with captions describing the attributes and a user editing request in natural language. 
        
        In a similar line of research, TediGAN~\cite{xia2021tedigan} proposes a novel GAN inversion technique that can map multi-modal information (i.e., texts, sketches, labels) into a common latent space of a pre-trained StyleGAN~\cite{stylegan2_karras20cvpr}. Then, using visual-linguistic similarity learning, it provides an interactive system where the user provides input textual guidelines and the system generates diverse images given the same input text, allowing the user to edit the appearance of different attributes interactively. However, for doing so, it relies on both annotated training data and the hierarchical structure of the StyleGAN architecture. More specifically, the authors of~\cite{xia2021tedigan} introduce and use the Multi-Modal CelebA-HQ, a curated dataset of face images, each having a high-quality segmentation mask, sketch, and descriptive text. Then, they use the hierarchical structure of StyleGAN's $\mathcal{W}$-space (i.e., style-mixing) to learn the text-image matching by mapping the image and text into the same joint embedding space. By contrast, our method is not limited to the task of face editing, is not tied to a specific GAN architecture (i.e., StyleGAN), and can be applied to any latent space.
        
        Finally, a work closely related to ours, StyleCLIP~\cite{patashnik2021styleclip}, utilizes a CLIP-based similarity loss in order to learn a StyleGAN-specific latent mapper network that infers a text-guided latent manipulation step for a given input image. To do so, StyleCLIP uses the CLIP text features of the given prompt as the target vector and tries to align it with the CLIP image features of the manipulated image. This leads to abrupt image manipulations and quickly leads to regions of low density and, thus, low image quality, as shown by the experimental results. By contrast, our method is architecture-agnostic (i.e., not tied to a specific GAN architecture, such as StyleGAN) and models non-linear paths driven by semantic dipoles in the text feature space that are being used for aligning the desired non-linear latent paths (Fig.~\ref{fig:summary}), arriving at longer and more disentangled traversals and preventing traveling to regions of low density, leading to images of higher quality with far better control of the generative factors.
    
    \begin{figure}[t]
        \centering
        \includegraphics[width=\textwidth]{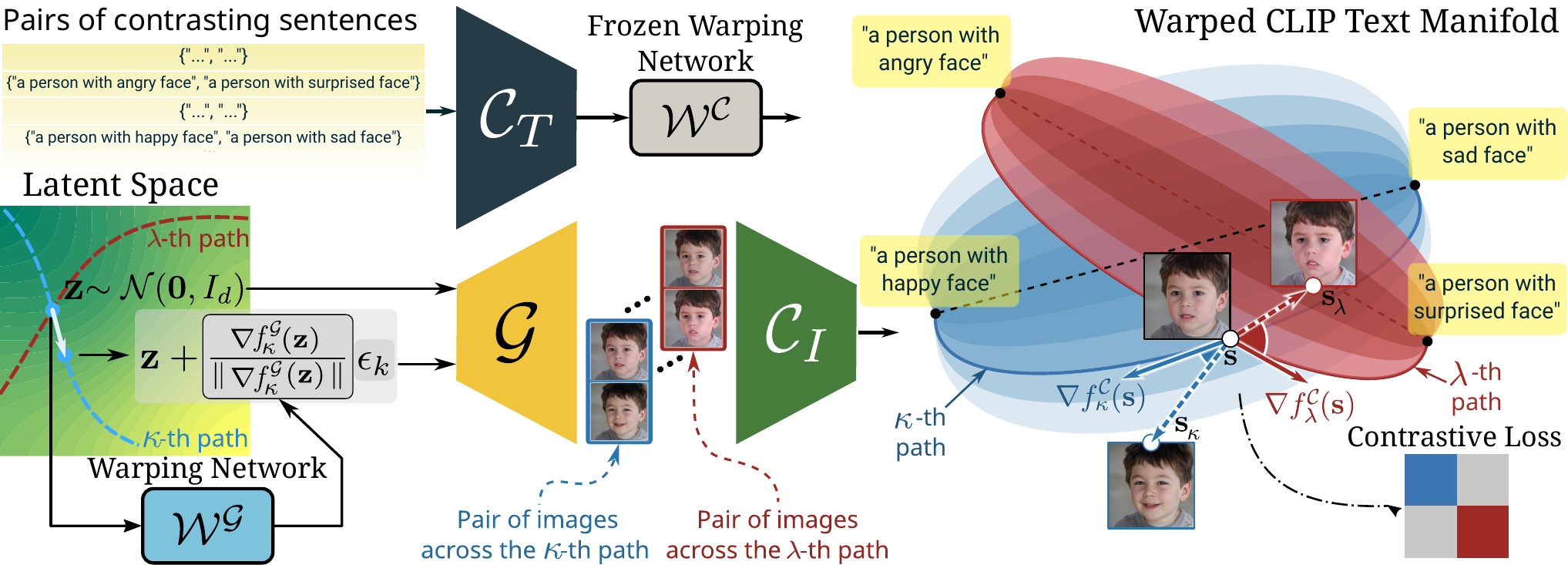}
        \caption{Overview of ContraCLIP -- The pre-trained CLIP~\cite{clip-radford2021icml} image-text space (given by the Image Encoder $\mathcal{C}_I$ and Text Encoder $\mathcal{C}_T$), warped due to \textit{semantic dipoles} of contrasting pairs of sentences given in natural language via the non-trainable warping network $\mathcal{W}^{\mathcal{C}}$, provides supervision to the optimisation of non-linear interpretable paths in the latent space of a pre-trained GAN $\mathcal{G}$ via the trainable warping network $\mathcal{W}^{\mathcal{G}}$.}
        \label{fig:overview}
    \end{figure}

\section{Proposed Method}\label{sec:proposed_method}

    Fig.~\ref{fig:overview} gives an overview of the proposed method for learning non-linear paths in the latent space of a pre-trained GAN, driven by contrasting semantic dipoles given in natural language. To do so, we define pairs of sentences that convey contrasting meanings and express the limits of the interpretation that we require by the optimised latent paths to encode. Each such pair corresponds to one trainable path in the GAN latent space. Given $K$ semantic dipoles, we first represent them in the CLIP text space and then use them as the centres of the RBF-based warping functions $f_1^{\mathcal{C}},\ldots,f_{\kappa}^{\mathcal{C}},\ldots,f_K^{\mathcal{C}}$. This endows us with $K$ distinct vector fields (the functions' gradients $\nabla f_{\kappa}^{\mathcal{C}}$), which provide end-to-end directional paths from the one pole/sentence to the other and can be used as our supervisory signal that guides the trainable latent paths, for any given image and its transformed version along a certain latent path. In contrast to the standard text-image similarity (e.g.,~\cite{patashnik2021styleclip}), we do not try to blindly align the transformed image vectors with the text feature of the desired end. Instead, we decouple the supervisory text features from the origin of axes, by taking the gradient of the warping at any given point (i.e., image generated by the GAN and represented in the joint image-text CLIP space). By doing so, we establish only local relations between the text and the image features. This leads to smooth non-linear text paths that allow for mild transition between the given point and the desired end. 
    
    The rest of this section is organized as follows: a) in Sect.~\ref{sec:warpedganspace} we present our method for learning non-linear paths in the latent space of a pre-trained GAN using a modified version of the publicly available implementation of WGS~\cite{warpedganspace2021iccv} -- this is illustrated in the bottom-left part of Fig.~\ref{fig:overview}, b) in Sect.~\ref{sec:text_paths} we present the proposed method for modeling non-linear paths in the CLIP text space driven by semantic dipoles so as to derive a supervisory signal for learning the WGS latent paths -- this is illustrated in the right part of Fig.~\ref{fig:overview}, c) in Sect.~\ref{sec:loss} we discuss the proposed contrastive loss -- this is illustrated in the bottom-right part of Fig.~\ref{fig:overview} and in Fig.~\ref{fig:standard_vs_proposed_sim}.
    
    \subsection{Non-linear latent paths on the GAN's latent space}\label{sec:warpedganspace}
        In order to learn non-linear paths in the latent space of a pre-trained GAN, we build on 
        WGS~\cite{warpedganspace2021iccv}. WGS optimises a set of $K$ warpings, $f_1^{\mathcal{G}},\ldots,f_K^{\mathcal{G}}$, in the GAN's latent space, each parametrized by a set of RBF-based latent space warping functions, as follows
        \begin{equation}\label{eq:wgs_f}
            f_\kappa^{\mathcal{G}}(\mathbf{z})=
            \sum_{i=1}^{N} 
            \left(
            \exp\left(-\gamma_i^\kappa \lVert\mathbf{z}-\mathbf{q}_i^\kappa\rVert^2\right)
            -
            \exp\left(-\gamma_i^\kappa \lVert\mathbf{z}+\mathbf{q}_i^\kappa\rVert^2\right)
            \right)
            ,\:\kappa\in\{1,\ldots,K\},
        \end{equation}
        where $\mathbf{q}_i^\kappa$ and $\gamma_i^\kappa$ denote the support vectors and the scaling parameters, respectively. Then, each $f_\kappa^{\mathcal{G}}$ gives rise to a family of non-linear paths via $\nabla f_\kappa^{\mathcal{G}}(\mathbf{z})$.
        
        In~\cite{warpedganspace2021iccv}, the trainable parameters of the set of RBFs (i.e., $\mathbf{q}_i^\kappa$ and $\gamma_i^\kappa$) are optimised so that the images that are generated by codes along different paths, are easily distinguishable by a discriminator network. By contrast, in this work, the training objective is that the differences in the generated images are aligned with paths connecting the semantic dipoles expressed in natural language. This is schematically shown in Fig.~\ref{fig:overview} and will be detailed in the following section. 
    
    \subsection{Non-linear paths on the CLIP text space using semantic dipoles}\label{sec:text_paths}
    
        \begin{figure}[t]
            \centering
            \includegraphics[width=\textwidth]{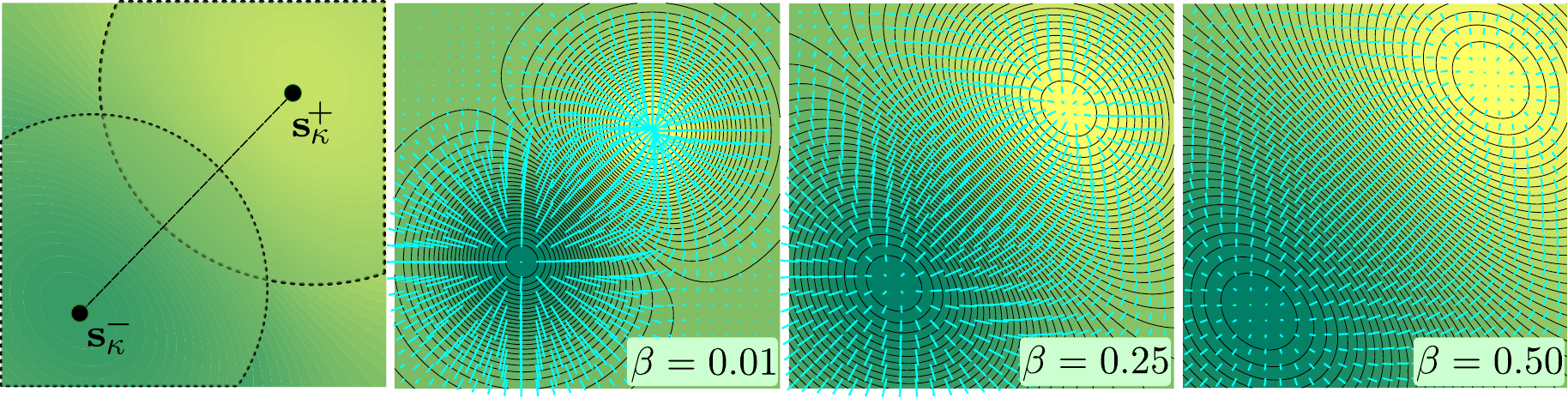}
            \caption{Estimation of the $\gamma$ parameter for the dipole of sentence $(\mathbf{s}_\kappa^-,\mathbf{s}_\kappa^+)$. We require that the two RBFs overlap by a factor $\beta\in(0,1)$; i.e., $\exp\left(-\gamma_\kappa\lVert\mathbf{s}_\kappa^+-\mathbf{s}_\kappa^-\rVert^2\right)=\beta$, or $\gamma_\kappa=-\frac{\log(\beta)}{\lVert\mathbf{s}_\kappa^+-\mathbf{s}_\kappa^-\rVert^2}$. $\beta$ governs the non-linearity of the formed paths.}
            \label{fig:dipole_betas}
        \end{figure}
    
        \begin{table}[t]
            \centering
            \caption{Example corpus of pairs of contrasting sentences in natural language. Each pair expresses the starting and the ending points of the latent path that needs to be learnt. Each such pair, when represented in the CLIP text space, forms a \textit{semantic dipole}, which via (\ref{eq:f_t}), is used to warp the text space and form non-linear directional paths from one pole to the other (see also Fig.~\ref{fig:overview}).}
            \begin{tabular}{cc}
            \hline
            \multicolumn{2}{c}{$\mathbf{s}^-$ $\longrightarrow$ $\mathbf{s}^+$} \\ \hline\hline
            \multicolumn{2}{c}{``a picture of a face in neutral expression.'' $\longrightarrow$ ``a picture of a smiling face.''}             \\ 
            \multicolumn{2}{c}{``a picture of a person with open eyes.'' $\longrightarrow$ ``a picture of a person with closed eyes.''}       \\ 
            \multicolumn{2}{c}{``a picture of a young person.'' $\longrightarrow$ ``a picture of an old person.''}                            \\ 
            \multicolumn{2}{c}{``a picture of a male face.'' $\longrightarrow$ ``a picture of a female face.''}                               \\ 
            \multicolumn{2}{c}{``a picture of a person with happy face.'' $\longrightarrow$ ``a picture of a person with fearful face.''}     \\ \hline
            \end{tabular}
            \label{tab:corpus}
        \end{table}
        
        Given a set (corpus) of pairs of sentences, each pair conveying contrasting meanings, we first use the pre-trained CLIP Text Encoder ($\mathcal{C}_T$) in order to represent them in the text space $\mathcal{S}\subset\mathbb{R}^{512}$. That is, each pair of sentences in natural language is represented by a pair of vectors in the CLIP text space forming a dipole. An example of such corpus of sentences is given in Tab.~\ref{tab:corpus}.
        
        Inspired by~\cite{warpedganspace2021iccv}, we propose the warping of the CLIP text space using RBF-based warping functions, each being the sum of two opposite RBFs centred at the text feature representations of the above dipoles. Each such warping of the text space endows it, via its gradient, with a family of non-linear curves as illustrated in Figs.~\ref{fig:summary},\ref{fig:overview}. That is, for any embedding $\mathbf{s}\in\mathcal{S}$, the gradient of the warping due to a certain semantic dipole gives a local direction vector, following which will eventually lead to one of the semantic poles. In other words, a warping function ensures that starting from any point in the text space, following the gradient of the function, one can --at will-- arrive at any of the two poles.
        
        More formally, let $\{(\mathbf{s}_\kappa^-,\mathbf{s}_\kappa^+)\colon\mathbf{s}_\kappa^{\pm}\in\mathcal{S}, \kappa=1,\ldots,K\}$ be the corpus of $K$ pairs of sentences represented in the $512$-dimensional CLIP text space $\mathcal{S}$. For the $\kappa$-th pair, $(\mathbf{s}_\kappa^-,\mathbf{s}_\kappa^+)$, we define the following warping function
        \begin{equation}\label{eq:f_t}
            f_\kappa^{\mathcal{C}}(\mathbf{s}) = 
            \exp\left(-\gamma_\kappa \lVert\mathbf{s}-\mathbf{s}_\kappa^+\rVert^2\right)-
            \exp\left(-\gamma_\kappa \lVert\mathbf{s}-\mathbf{s}_\kappa^-\rVert^2\right),
        \end{equation}
        where $\gamma_\kappa\in\mathbb{R}_+$ denotes the scale of each of the poles in the dipole and governs the non-linearity of the resulting paths. In Fig.~\ref{fig:dipole_betas} we illustrate the proposed approach for setting the $\gamma$ parameters of each pair of sentences of the given corpus. By requiring that the RBFs of each dipole will overlap by a given factor $\beta$, we arrive at warpings of the text feature space that have non-zero gradient and, thus, allow for traversing the space continuously from any given point to the desired end of the path. We note that $\gamma_\kappa$ depends on the relative positions of the sentences $\mathbf{s}_\kappa^-$ and $\mathbf{s}_\kappa^+$ and, thus, are in general different for each pair. The gradient of each warping function is given analytically as
        \begin{equation}\label{eq:nabla_f_t}
            \nabla f_\kappa^{\mathcal{C}}(\mathbf{s}) =
            -2\gamma_\kappa
            \left(
            \exp\left(-\gamma_\kappa\lVert\mathbf{s}-\mathbf{s}_\kappa^+\rVert^2\right)\left(\mathbf{s}-\mathbf{s}_\kappa^+\right)
            -
            \exp\left(-\gamma_\kappa\lVert\mathbf{s}-\mathbf{s}_\kappa^-\rVert^2\right)\left(\mathbf{s}-\mathbf{s}_\kappa^-\right)
            \right).
        \end{equation}
    
    \subsection{Proposed contrastive similarity loss}\label{sec:loss}
        In this section we describe our objective for learning the trainable parameters of the set of RBFs, i.e., the support vectors $\mathbf{q}_i^\kappa$ and the scaling parameters $\gamma_i^\kappa$ in (\ref{eq:wgs_f}). We propose a contrastive loss term that imposes alignment between the vector that expresses the difference between the manipulated and the original image, and the local text vector (tangent to the non-linear curve) that is obtained in the CLIP text space by the non-trainable warping network $\mathcal{W}^{\mathcal{C}}$ (Fig.~\ref{fig:overview}).
        
        Formally, given a mini-batch of $N$ images generated by the GAN generator and represented in the joint image-text CLIP space, the loss term introduced by the pair of images due to the $\kappa$-th latent path, i.e., by the original image $\mathbf{s}$ and the manipulated image $\mathbf{s}_{\kappa}$, is given by
        \begin{equation}\label{eq:loss}
            \ell_{\kappa}=
            -\log\frac{\exp\left(\sfrac{p_{\kappa,\kappa}}{\tau}\right)}{\sum_{t=1}^{N}\exp\left(\sfrac{p_{\kappa,t}}{\tau}\right)},
        \end{equation}
        where $p_{\kappa,t}=\cos\left(\nabla f_{\kappa}^{\mathcal{C}}(\mathbf{s}),\mathbf{s}_t-\mathbf{s}\right)$ and $\tau$ denotes the temperature parameter. It is worth noting that the only trainable module of the proposed method is the set of the support vectors that warp the latent space of GAN (i.e., the trainable Warping Network $\mathcal{W}^{\mathcal{G}}$ shown in the bottom left part of Fig.~\ref{fig:overview}).

\section{Experiments}\label{sec:experiments}
    
    In this section we will present the experimental evaluation of the proposed method and provide comparisons with state-of-the-art methods. We will first briefly present the types of GAN generators and the datasets that we use in Sect.~\ref{sec:gans_and_datasets}. We will then present ablation studies on various components of the proposed method in Sect.~\ref{sec:ablation_studies}, and we will show that using the semantic dipoles to warp the CLIP text space in order to find non-linear text paths is crucial with respect to the quality of the discovered latent paths (both in terms of image quality and disentanglement). More precisely, as shown in Figs.~\ref{fig:proggan_smiling},\ref{fig:proggan_makeup},\ref{fig:stylegan_expressions}, replacing the non-linear text paths with linear ones (e.g., the vector that connects the two poles) leads to significant deterioration of the discovered latent paths, both in terms of image quality and disentanglement, due to the abrupt image manipulations induced by the linear text directions. By contrast, non-linear text paths lead to smooth and more disentangled interpretable latent paths, generating images of higher quality and preserving attributes based on the given semantic dipoles. Moreover, due to the capability of traveling very long traversals in the latent space, using the proposed non-linear text paths leads to latent paths that exhibit far better control of the generative factors (Fig.~\ref{fig:ablation_traversal_lengths}). Finally, in Sect.~\ref{sec:soa_comparisons}, we will qualitatively and quantitatively show that the proposed method produces continuous and more disentangled latent path in comparison to state-of-the-art StyleCLIP~\cite{patashnik2021styleclip}, by preserving the identity (ID) significantly better, while exhibiting much lower correlation to facial attributes that are irrelevant to the given semantic dipoles (Tab.~\ref{tab:soa_dis}). We note that, in all experiments, in order to measure the ID preservation between the original image of each traversal sequence, i.e., the central image of the generated sequences across the various latent paths, and each of the rest on the path, we used the ID similarity score (i.e., a number in $[0,1]$) provided by ArcFace~\cite{deng2019arcface}. Finally, in order to measure the disentanglement of the discovered paths, we used the disentanglement metric proposed in~\cite{warpedganspace2021iccv}.
    
    \subsection{Pre-trained GAN generators and datasets}\label{sec:gans_and_datasets}
        In order to show how general our proposed approach is, we evaluate it using pre-trained GANs of different architectures, namely: a) ProgGAN~\cite{proggan_karras18iclr} trained on CelebA-HQ~\cite{celeba_liu15iccv}, and b) StyleGAN2~\cite{stylegan2_karras20cvpr} ($\mathcal{W}$-space) trained on FFHQ~\cite{stylegan2_karras20cvpr}. Additional experiments on StyleGAN2 trained on AFHQ Cats~\cite{choi2020stargan}, AFHQ Dogs~\cite{choi2020stargan}, and LSUN Cars~\cite{yu2015lsun} are given in the supplementary material.
    
    \subsection{Ablation studies}\label{sec:ablation_studies}
        In this section we will present our ablation studies on the length of the latent traversals and the type of text-image similarity measure used by our method. We also conducted ablation studies to show the robustness to the values of the $\beta$ parameter (Sect.~\ref{sec:text_paths}) and the temperature of the adopted contrastive loss (Sect.~\ref{sec:loss}), which we include in the supplementary material.
        
        \begin{figure}[t!]
            \centering
            \includegraphics[width=\textwidth]{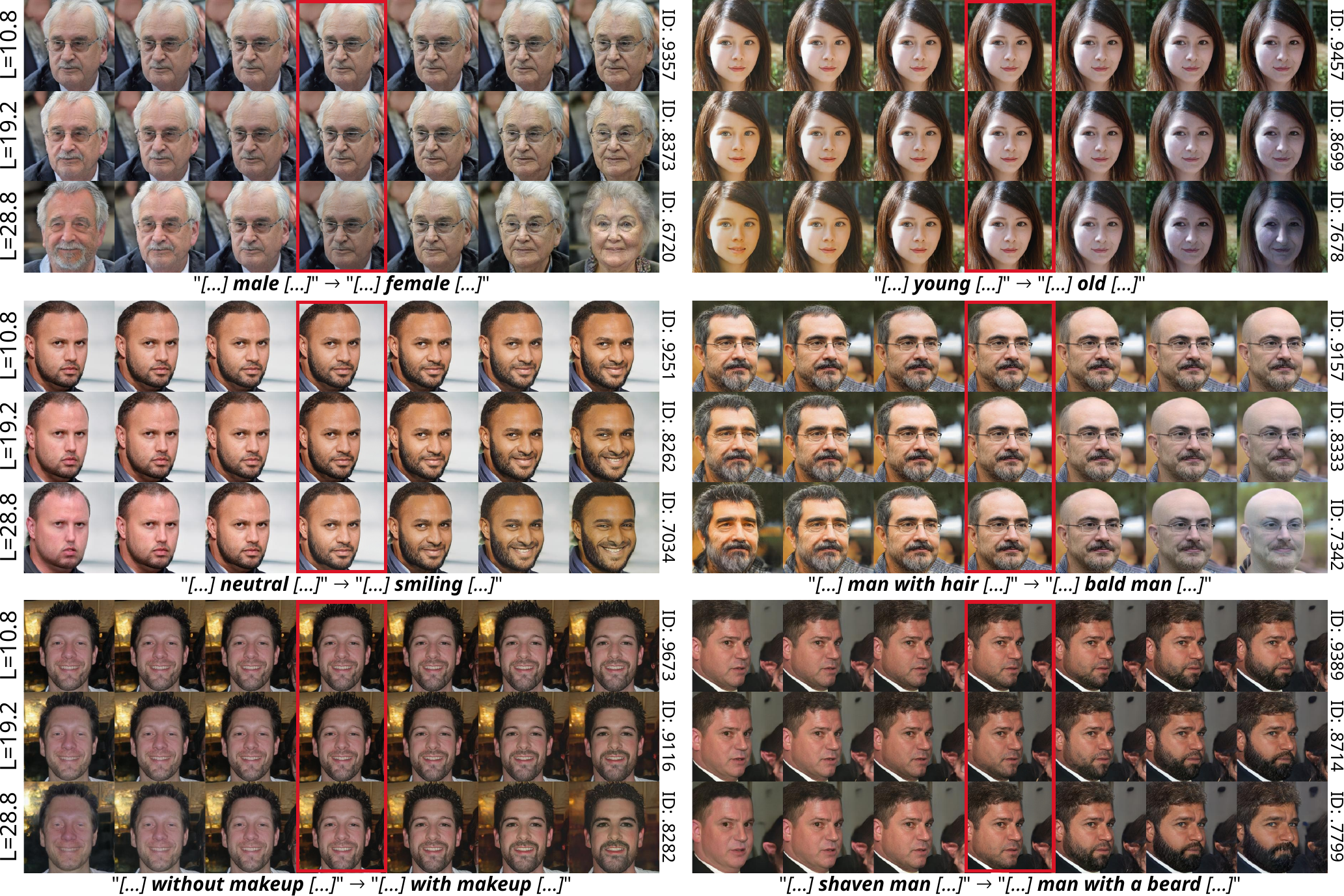}
            \caption{Ablation study on latent traversal length $L\in\{10.8,19.2,28.8\}$.}
            \label{fig:ablation_traversal_lengths}
        \end{figure}
        
        \begin{figure}[t!]
            \centering
            \includegraphics[width=\textwidth]{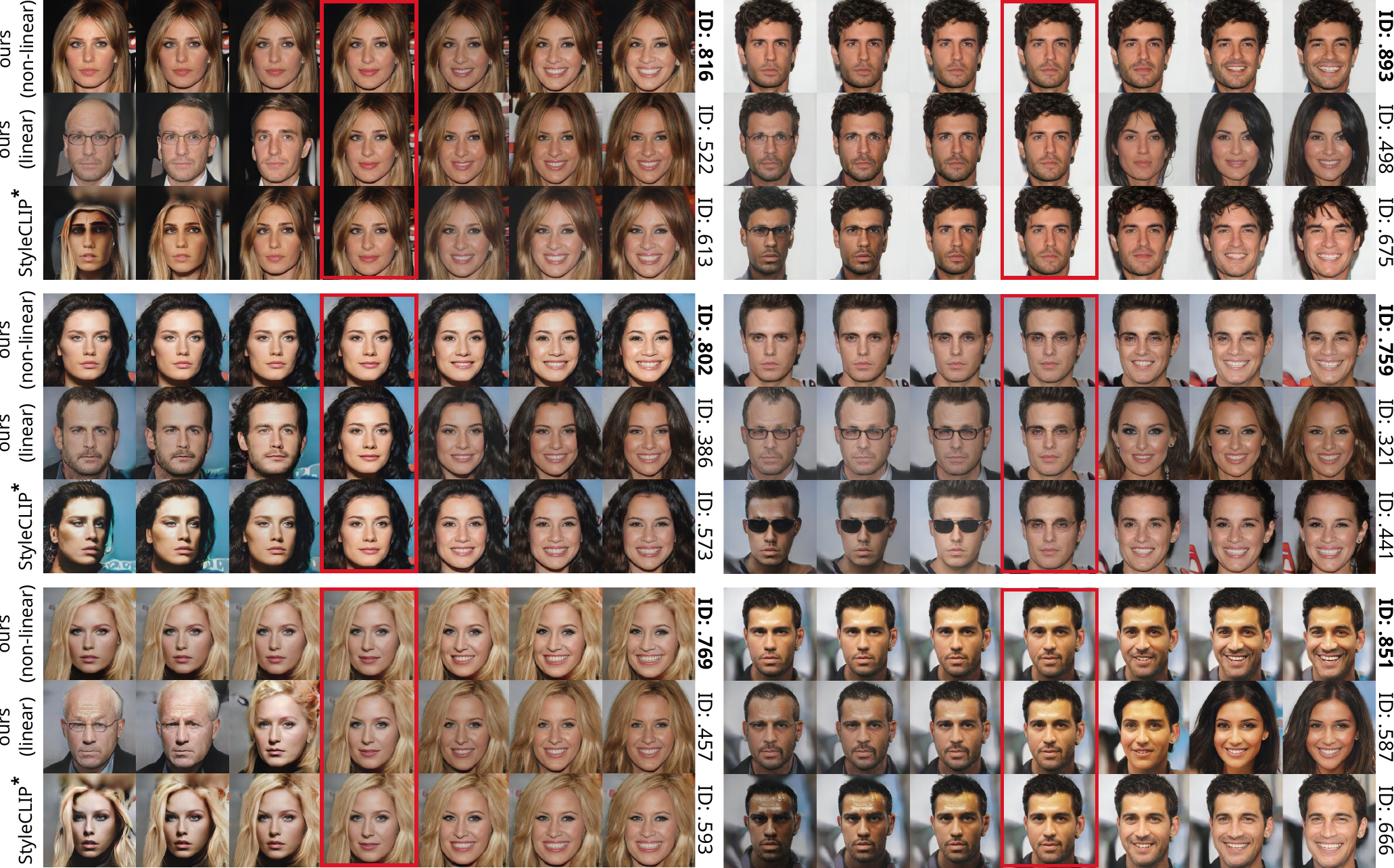}
            \caption{Ablation study on the use of non-linear versus linear text paths for the semantic dipole ``\textit{a face in neutral expression.}'' $\rightarrow$ ``\textit{a smiling face.}'' (ProgGAN).}
            \label{fig:proggan_smiling}
        \end{figure}
        
        \begin{figure}[t!]
            \centering
            \includegraphics[width=\textwidth]{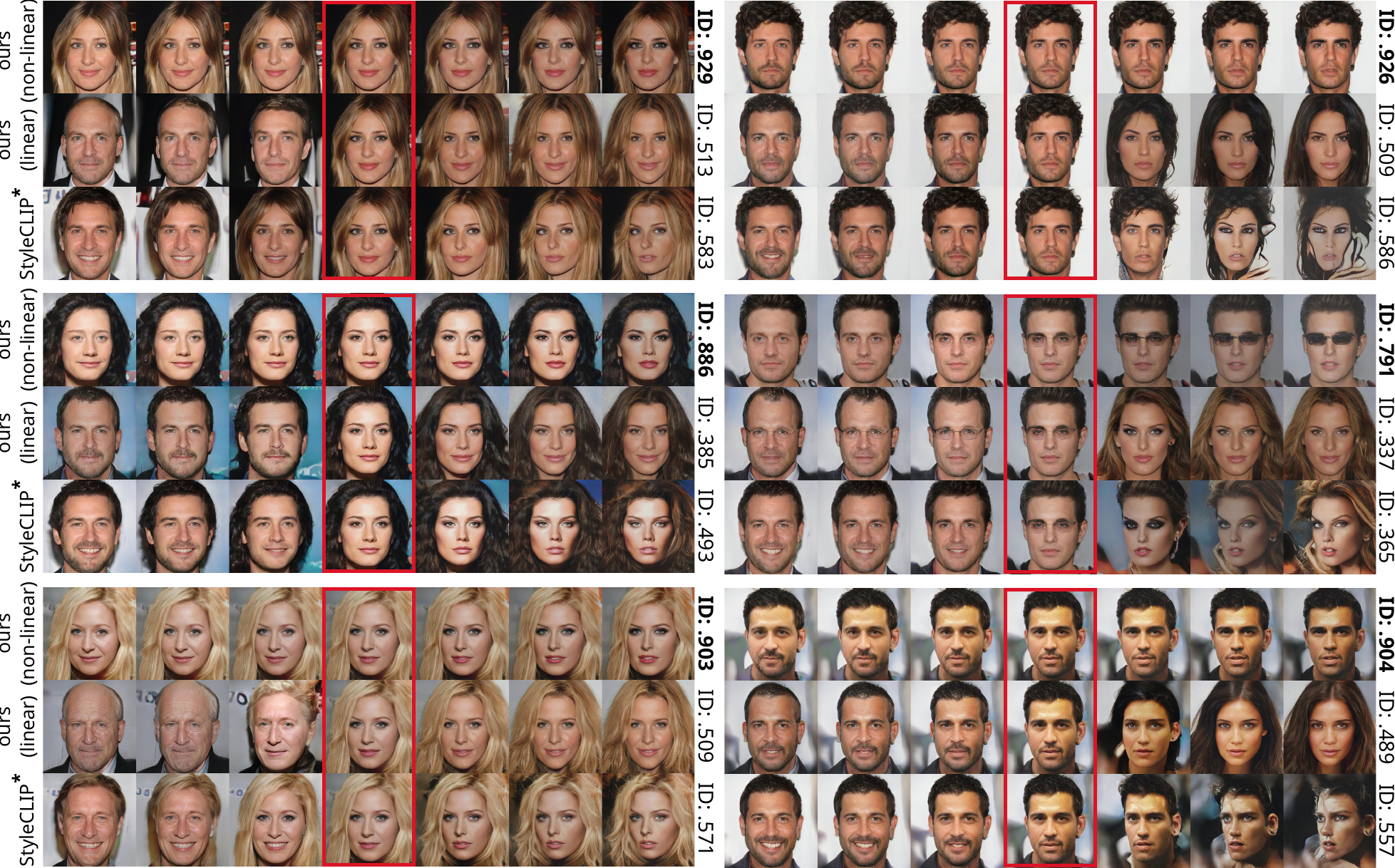}
            \caption{Ablation study on the use of non-linear versus linear text paths for the semantic dipole ``\textit{a face without makeup.}'' $\rightarrow$ ``\textit{a face with makeup.}'' (ProgGAN).}
            \label{fig:proggan_makeup}
        \end{figure}
        
        \begin{figure}[t!]
            \centering
            \includegraphics[width=\textwidth]{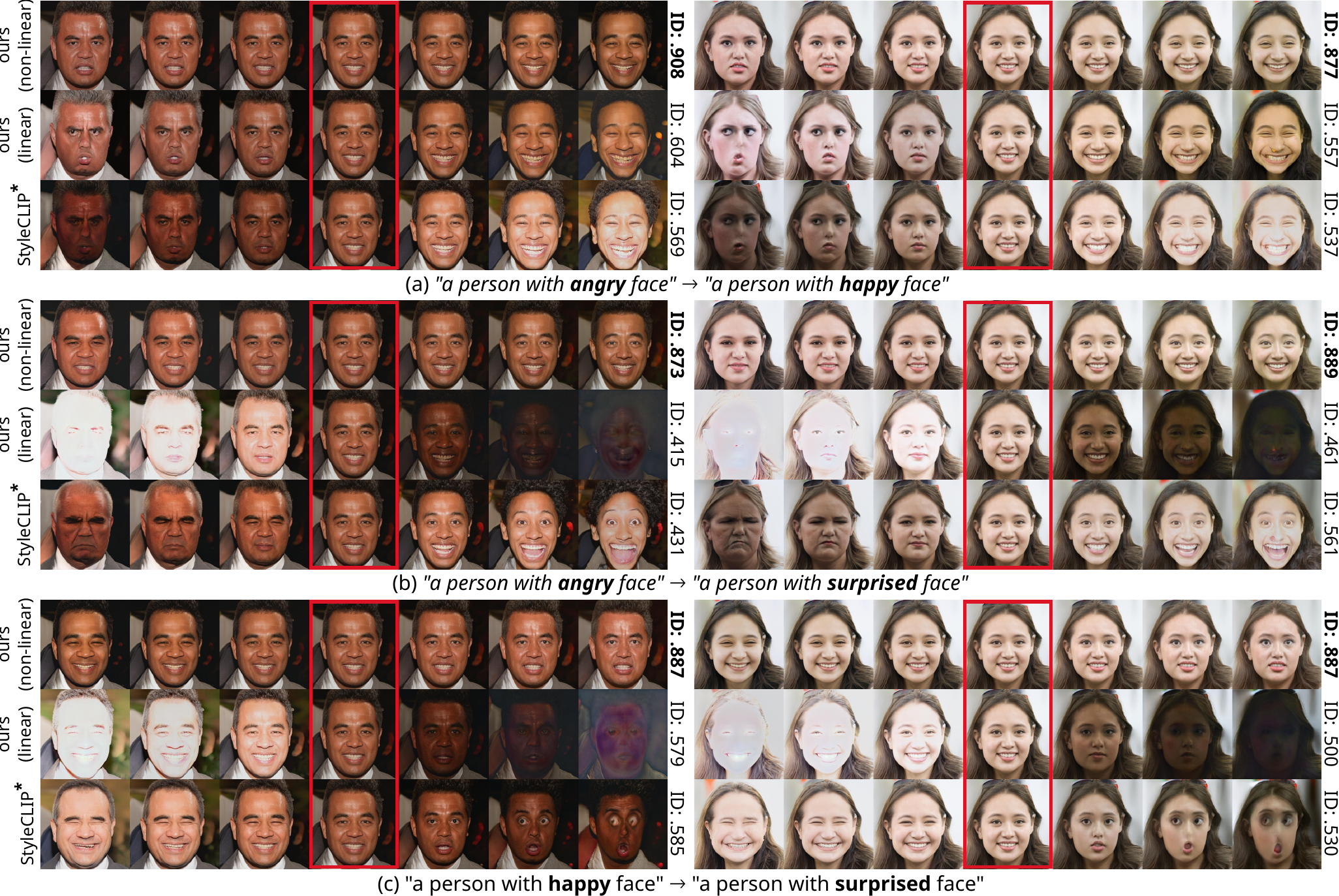}
            \caption{Ablation study on the use of non-linear vs. linear text paths (StyleGAN2).}
            \label{fig:stylegan_expressions}
        \end{figure}
        
        \subsubsection{Latent traversal length ablation}
            In order to explore the limitations of the proposed method with respect to the length of the traversal along the discovered latent paths, we conducted the following ablation study. We calculated the latent traversals for a set of latent codes (in StyleGAN2's $\mathcal{W}$-space) for a various number of steps, leading to latent traversals of length $L\in\{10.8,19.2,28.8\}$. We show the results in Fig.~\ref{fig:ablation_traversal_lengths}. Surprisingly, the proposed non-linear text paths continue to semantically extend towards each end of the dipole without arriving at low-density areas on the latent space and, thus, low-quality generations. It is worth noting that by using linear text paths in the CLIP space (e.g., StyleCLIP~\cite{patashnik2021styleclip}), we arrive at heavy artifacts or extreme deformations of the images even after traveling for relatively few ($L\approx11$) units of length (as will be shown later in this section), while in our case we can travel for $L\approx29$ without introducing extreme artifacts, and at the same time we semantically approach even closer to the ends of the dipoles. This leads to better control over the discovered generative factors and prevents traveling to regions of low density.
        
        \subsubsection{Non-linear vs linear text paths}
            In order to assess the effectiveness of warping the CLIP text space and modeling non-linear path on it, we conducted an ablation study on the similarity criterion that the proposed method can employ. First, we considered linear text paths formed as the difference of dipoles ends -- this is denoted as ``ours (linear)''. Next, we considered the standard text-similarity calculation approach when a single text prompt is given (for fair comparisons, we learned latent paths for both ends of the semantic dipole separately). This approach is illustrated in the left part of Fig.~\ref{fig:standard_vs_proposed_sim} and since it is adopted by StyleCLIP~\cite{patashnik2021styleclip} we denote this as ``StyleCLIP\textsuperscript{\textasteriskcentered}''. We show the results for ProgGAN in Figs.~\ref{fig:proggan_smiling},~\ref{fig:proggan_makeup} and for StyleGAN2 in Figs.~\ref{fig:stylegan_expressions}. It is clear that adopting a similarity calculation approach that involves linear text directions leads to abrupt manipulations of the corresponding image, usually causing severe artifacts, or altering entirely the identity and/or facial attributes depicted on it. By contrast, the proposed non-linear text paths lead to latent traversals that are smooth, effective, and far more disentangled. Consequently, adopting a semantic dipole (instead of single text prompts) is not enough without warping the text space in order to obtain non-linear paths between the centres of the dipole.

            \begin{figure}[t!]
                \centering
                \includegraphics[width=\textwidth]{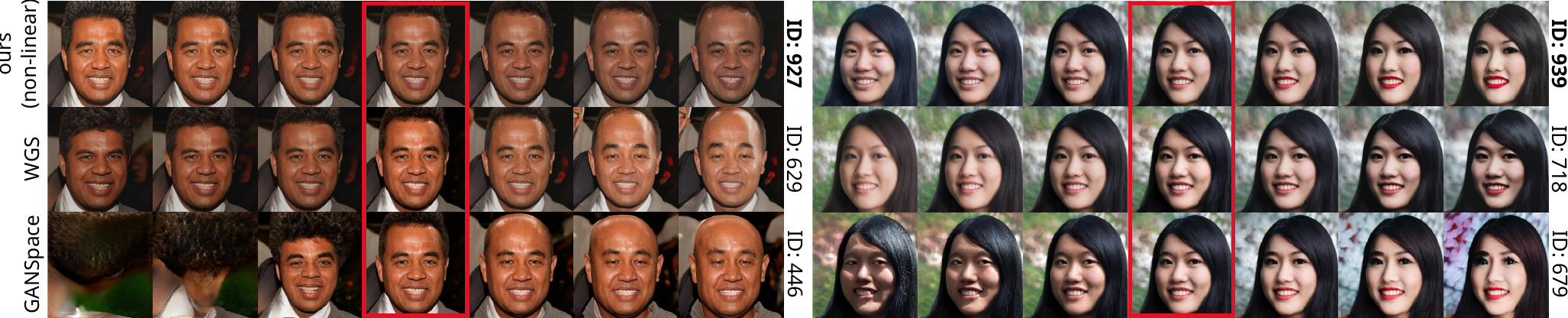}
                \caption{Comparison of the proposed method with GANSpace~\cite{ganspace2020harkonen} and WGS~\cite{warpedganspace2021iccv}.}
                \label{fig:comp_soa_ganspace_warpedganspace}
            \end{figure}
                
            \begin{figure}[t!]
                \centering
                \includegraphics[width=\textwidth]{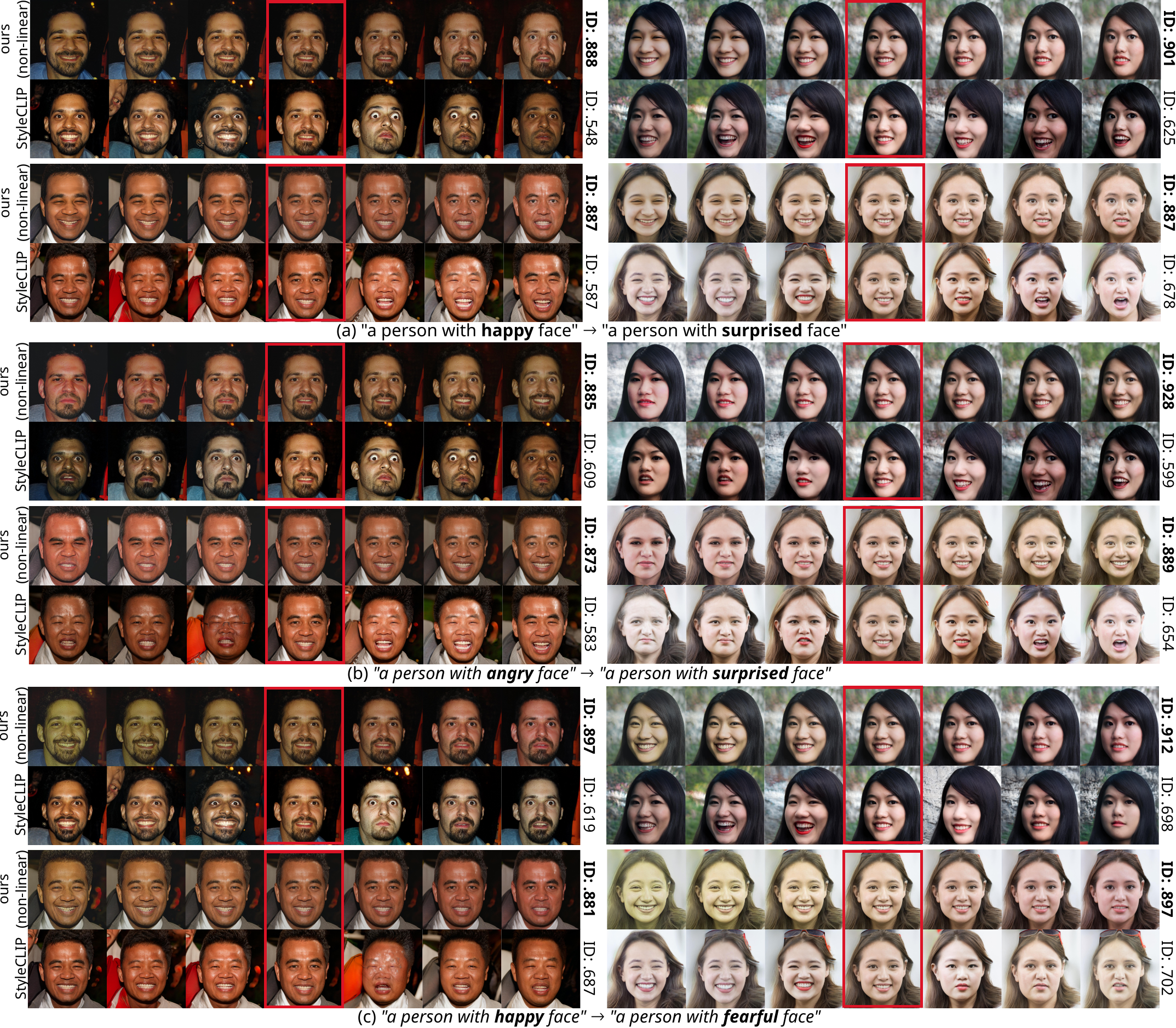}
                \caption{Comparison of the proposed method with StyleCLIP~\cite{patashnik2021styleclip}.}
                \label{fig:comp_soa_styleclip}
            \end{figure}
        
            \begin{figure}[t!]
                \centering
                \includegraphics[width=\textwidth]{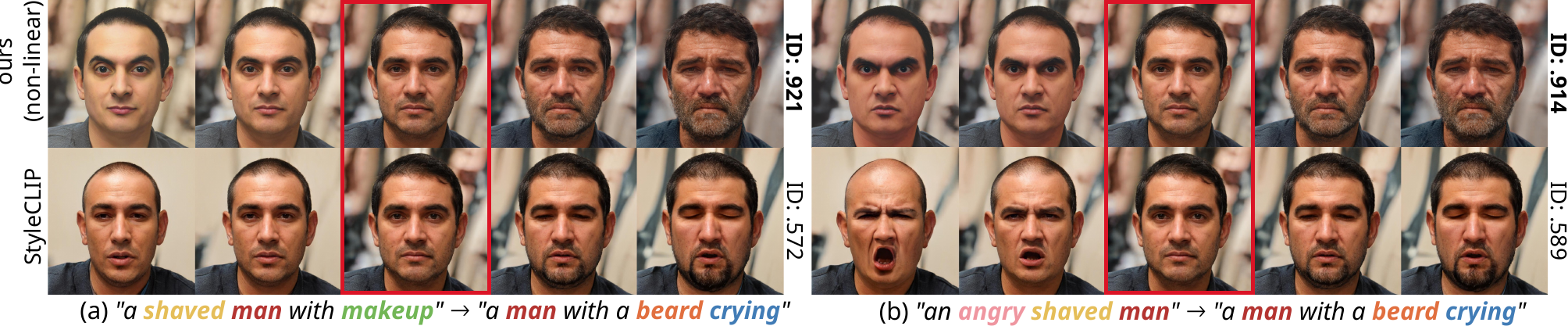}
                \caption{Comparison of the proposed method with StyleCLIP~\cite{patashnik2021styleclip} for more complex and challenging semantic dipoles.}
                \label{fig:comp_soa_complex}
            \end{figure}

    \begin{table}[t]
        \centering
        \caption{Comparison of the proposed ContraCLIP to state of the art methods in terms of the disentanglement metric proposed in~\cite{warpedganspace2021iccv}.}
        \begin{tabular}{l||c|ccccc}
                           & ID ($\uparrow$) & Age ($\downarrow$) & Gender ($\downarrow$) & Skin ($\downarrow$) & Hair ($\downarrow$) & Beard ($\downarrow$) \\ \hline\hline
            GANSpace~\cite{ganspace2020harkonen}         & .23 & .95 & .63 & .57 & .89 & .33 \\
            WarpedGANSpace~\cite{warpedganspace2021iccv} & .29 & .82 & .39 & .78 & .94 & .38 \\
            StyleCLIP~\cite{patashnik2021styleclip}      & .35 & .73 & .55 & .42 & .93 & .41 \\ \hline
            \textbf{ContraCLIP (Ours)}  & \textbf{.83} & \textbf{.12} & \textbf{.06} & \textbf{.30} & \textbf{.26} & \textbf{.14} \\ \hline
        \end{tabular}
        \label{tab:soa_dis}
    \end{table}

    \subsection{Comparison to state-of-the-art}\label{sec:soa_comparisons}

        In this section, we first compare the proposed method, which discovers non-linear paths in the GAN's latent space and may adopt non-linear or linear paths in the text space, with GANSpace~\cite{ganspace2020harkonen} and WGS~\cite{warpedganspace2021iccv}, which discover linear or non-linear paths, respectively, in the GAN latent space in an unsupervised manner. In Fig.~\ref{fig:comp_soa_ganspace_warpedganspace} we show that using text guidelines leads to more disentangled interpretable paths (e.g., by preserving certain attributes such as hair style and facial expressions) and far less artifacts, without the need of labeling them using manual annotation~\cite{ganspace2020harkonen} or certain pre-trained detectors~\cite{warpedganspace2021iccv}.
        
        In Figs.~\ref{fig:comp_soa_styleclip},\ref{fig:comp_soa_complex} we compare the proposed method with the state-of-the-art vision-language StyleCLIP~\cite{patashnik2021styleclip}. We observe that our method produces consistent and more disentangled latent paths (e.g., preserving certain attributes -- see also Tab.~\ref{tab:soa_dis}), which continuously traverses the latent space from a given semantic region to another, by preserving at the same time the identity significantly better than~\cite{patashnik2021styleclip} as is clear by the ID similarity values at the right hand side of each row. By contrast, StyleCLIP~\cite{patashnik2021styleclip} leads to inconsistent traversals that do not progress in a continuous manner and arrives at more unnatural generations (see also the more complex semantic dipoles shown in Fig.~\ref{fig:comp_soa_complex}).
        
        Finally, in Tab.~\ref{tab:soa_dis} we show state-of-the-art results using the quantitative disentanglement metric proposed in~\cite{warpedganspace2021iccv}. The results are for the facial expressions transition experiment shown in Fig.~\ref{fig:comp_soa_styleclip}. In Tab.~\ref{tab:soa_dis} we show that in comparison to state-of-the-art methods (i.e., \cite{ganspace2020harkonen,warpedganspace2021iccv,patashnik2021styleclip}) we obtain much higher ID preservation and much lower correlation to five facial attributes that are irrelevant to facial expressions -- this indicates better disentanglement.

\section{Conclusions}\label{sec:conclusions}

    In this paper, we presented our method for discovering non-linear paths in the latent space of pre-trained GANs driven by semantic dipoles. We do so by defining a set of contrasting pairs of sentences in natural language that represented in the CLIP text space give rise, via RBF-based warping functions, to non-linear text paths for traversing it from one semantic pole to the other. By defining an objective that discovers paths in the latent space of GANs that generate changes along the desired paths in the vision-language embedding space, we provide an intuitive and effective way of controlling the underlying generating factors.

\begin{appendices}
In this supplementary material we will present a) additional ablation studies with respect to hyperparameters of the proposed method in Sect.~\ref{supp:ablation_params}, b) quantitative experimental results on the preservation of the identity between the original sampled images and the generated image sequences across the various learnt latent paths in Sect.~\ref{supp:id_preservation}, and c) additional experimental results on non-facial datasets in Sect.~\ref{supp:other_datasets}.

\section{Ablation studies with respect to the hypeparameters $\beta$ and $\tau$}\label{supp:ablation_params}
    In this section we will present additional ablation studies to show the robustness of the proposed method to the values of the $\beta$ parameter (see Sect.~\ref{sec:text_paths}) and the temperature of the adopted contrastive loss (see Sect.~\ref{sec:loss}).
    
    \subsection{$\beta$ parameter}
        As discussed in Sect.~\ref{sec:text_paths}, we introduce the parameter $\beta$ as a means of controlling the $\gamma$ parameter of each semantic dipole, or, in other words, a way of controlling the non-linearity of the curves that are induced by the warping of the CLIP text space due to function (\ref{eq:nabla_f_t}). In Fig.~\ref{fig:dipole_betas} we illustrate the proposed approach for setting the $\gamma$ parameters of each pair of sentences (semantic dipole) of a given corpus. More specifically, by requiring that the RBF centred on the one pole, evaluated on the other pole, has a value of $\beta\in(0,1)$, we arrive at warpings of the text feature space that have non-zero gradient and, thus, allow for traversing the space continuously from any given point to the desired end of the path. We note that $\gamma_t$ depends on the relative positions of the sentences $\mathbf{s}_t^-$ and $\mathbf{s}_t^+$ and, thus, are in general different for each pair. 
        
\begin{figure}[t!]
            \centering
            \includegraphics[width=0.95\textwidth]{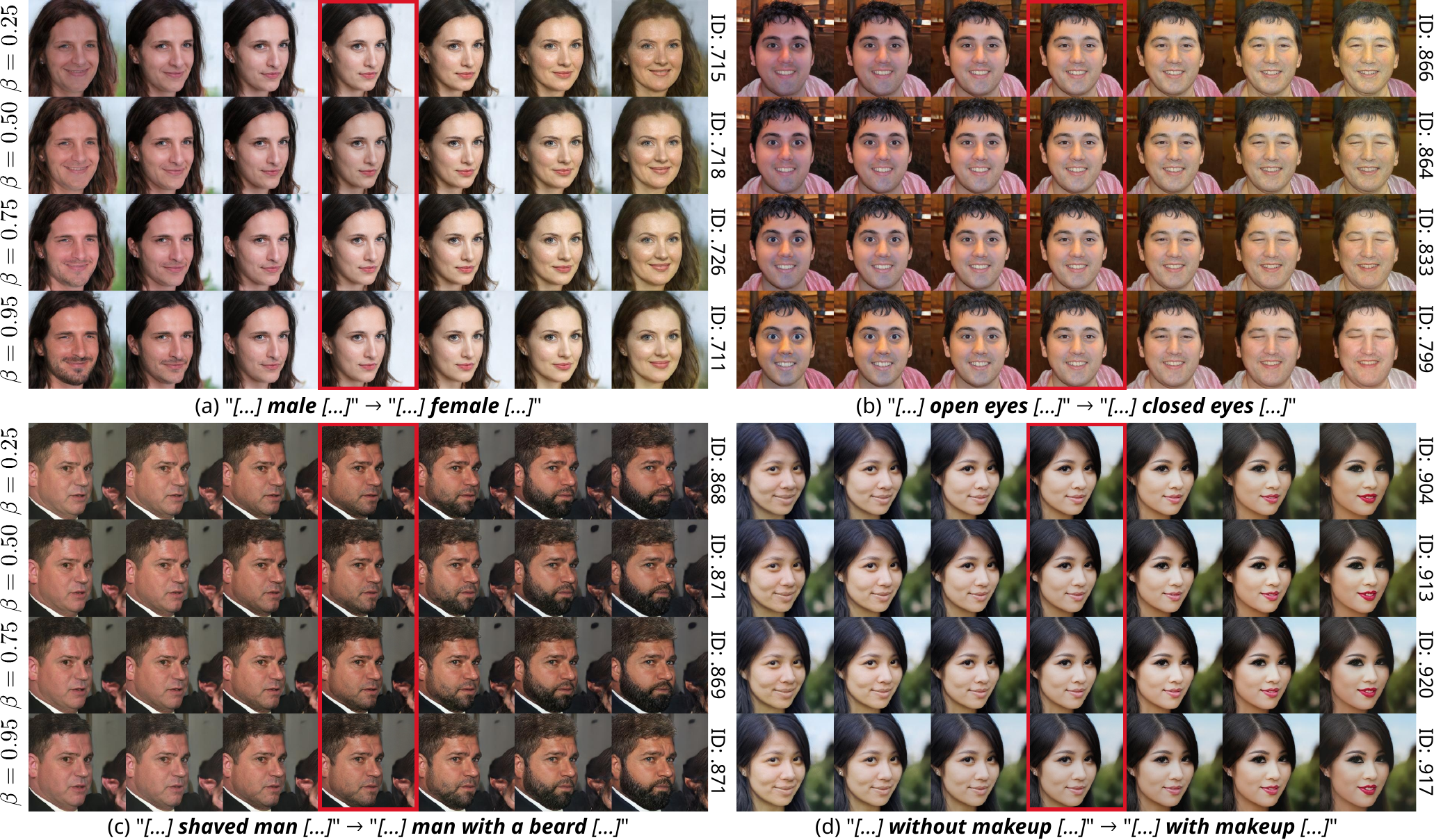}
            \caption{Ablation study on the $\beta$ parameter.}
            \label{fig:ablation_betas}
        \end{figure}

        A $\beta\to0$ leads to non-overlapping RBFs and thus greater flat regions around the poles and, thus, zero gradient (see the case of $\beta=0.01$ in Fig.~\ref{fig:dipole_betas}). In contrast, when $\beta\to1$ the RBFs largely overlap with each other by flattening the RBF ``bells''. We experimented with values of $\beta\in\{0.25, 0.75, 0.95\}$ and we show that the discovered latent paths are very similar to each other for the various choices of $\beta$ in this range, as shown in Fig.~\ref{fig:ablation_betas}. The visual differences one can observe are subtle in general, while the ID scores (i.e., an identity score for each image of the sequence that expresses the similarity between the original image -- central image of the sequence -- and each of the rest, using ArcFace~\cite{deng2019arcface}) are close to each other. Finally, it is worth noting that for smaller values of $\beta$, i.e., $\beta\in(0,0.2)$, the training process collapses since there are large flat regions around the poles which do not allow gradient to show to the desired direction.
    
    \subsection{Contrastive temperature parameter $\tau$}
\begin{figure}[t!]
            \centering
            \includegraphics[width=0.95\textwidth]{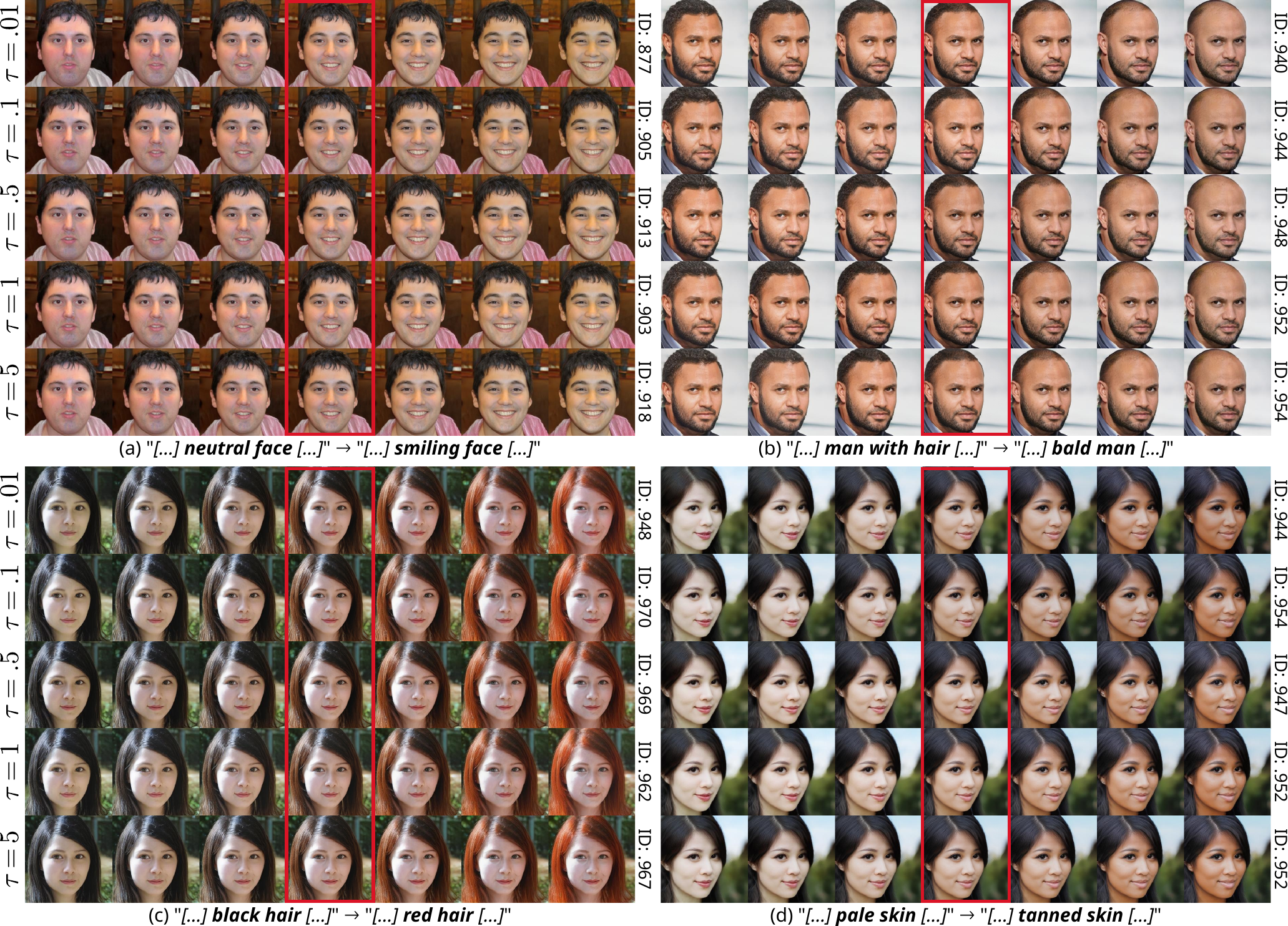}
            \caption{Ablation study on the contrastive loss temperature parameter $\tau$.}
            \label{fig:ablation_temperature}
        \end{figure}

        We also conducted a study on the temperature of the proposed contrastive loss (see Sect.~\ref{sec:loss}). In this section we will present our ablation study on the temperature $\tau$ and we will show that the proposed method is also robust with respect this parameter as shown in Fig.~\ref{fig:ablation_temperature}. We experimented with temperature values $\tau\in\{0.01,0.1,0.5,1.0,5.0\}$. Similarly to the previous section, we observe that the proposed method exhibits notable robustness with respect to the adopted temperature parameter.
        
\section{ID preservation}\label{supp:id_preservation}
    
    \begin{figure}[t!]
        \centering
        \includegraphics[width=0.95\textwidth]{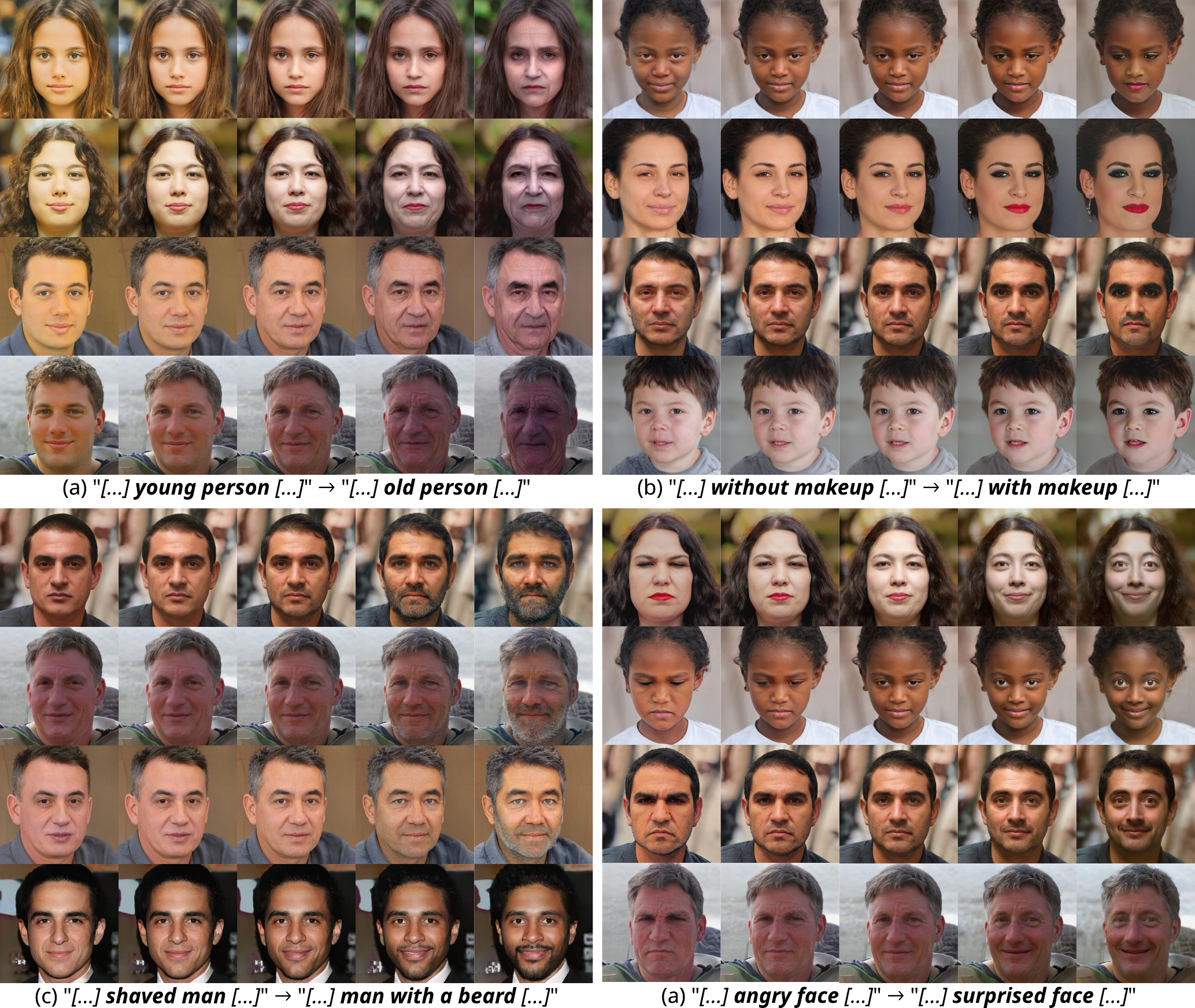}
        \caption{Examples of the interpretable paths found with the proposed method for the dipoles: (a) ``a picture of a young person.'' $\longrightarrow$ ``a picture of an old person.'', (b) ``a picture of a shaved man.'' $\longrightarrow$ ``a picture of a man with beard.'', and (c) ``a picture of a face without makeup.'' $\longrightarrow$ ``a picture of a face with makeup.''.}
        \label{fig:stylegan_various_corpus}
    \end{figure}
    
    As discussed in Sect.~\ref{sec:experiments}, for measuring the identity preservation between the original image of each traversal sequence, i.e., the central image of the generated sequences across the various latent paths, and each of the rest on the path, we used an ID similarity score (ID), i.e., a number in [0, 1], provided by ArcFace~\cite{deng2019arcface}.
    
    In this section, we present quantitative results for a set of 100 randomly chosen latent codes (i.e., original images) and for a set of non-linear latent paths optimised for the semantic dipoles shown in Table~\ref{tab:corpus}. Examples of such latent paths for various latent codes and four semantic dipoles are shown in Fig.~\ref{fig:stylegan_various_corpus}. In this section we report results for the proposed non-linear method -- denoted as ``ContraCLIP'' and one that is adopted by the state-of-the-art StyleCLIP~\cite{patashnik2021styleclip} -- denoted as ``StyleCLIP\textsuperscript{\textasteriskcentered}'', as discussed in Sect.~\ref{sec:ablation_studies}. In Table~\ref{tab:id_results} we show the results of ID preservation (averaged over the 100 latent codes). Clearly, the proposed non-linear text paths lead to far better preservation of the identity of the original images (original latent codes) compared to linear ones -- see also Figs.~\ref{fig:proggan_smiling},\ref{fig:proggan_makeup},\ref{fig:stylegan_expressions}.
    
    \begin{table}[t!]
        \centering
        \caption{Corpus of semantic dipoles (\textit{Facial Attributes}).}
        \begin{tabular}{cc}
        \hline
        Semantic \\ Dipole & $\mathbf{s}^-$ $\longrightarrow$ $\mathbf{s}^+$ \\ \hline\hline
        $D_1$ & ``a picture of a male face.'' $\longrightarrow$ ``a picture of a female face.'' \\
        $D_2$ & ``a picture of a young person.'' $\longrightarrow$ ``a picture of an old person.'' \\
        $D_3$ & ``a picture of a smiling face.'' $\longrightarrow$ ``a picture of a face in neutral expression.'' \\
        $D_4$ & ``a picture of a person with black hair.'' $\longrightarrow$ ``a picture of a person with red hair.'' \\
        $D_5$ & ``a picture of a man with hair.'' $\longrightarrow$ ``a picture of a bald man.'' \\
        $D_6$ & ``a picture of a shaved man.'' $\longrightarrow$ ``a picture of a man with a beard.'' \\
        $D_7$ & ``a picture of a face without makeup.'' $\longrightarrow$ ``a picture of a face with makeup.'' \\
        $D_8$ & ``a picture of a person with open eyes.'' $\longrightarrow$ ``a picture of a person with closed eyes.'' \\
        $D_9$ & ``a picture of a person with pale skin.'' $\longrightarrow$ ``a picture of a person with tanned skin.'' \\ 
        $D_{10}$ & ``a picture of an angry face.'' $\longrightarrow$ ``a picture of a surprised face.'' \\ \hline
        \end{tabular}
        \label{tab:corpus}
    \end{table}
    
    \begin{table}[t!]
        \centering
        \caption{ID preservation results.}
        \begin{tabular}{lcccccccccc}
        \multirow{2}{*}{}                      & \multicolumn{10}{c}{Semantic Dipole} \\ \cline{2-11} 
                                               & $D_{1}$ & $D_{2}$ & $D_{3}$ & $D_{4}$ & $D_{5}$ & $D_{6}$ & $D_{7}$ & $D_{8}$ & $D_{9}$ & $D_{10}$ \\ \hline\hline
        \multicolumn{1}{l|}{StyleCLIP\textsuperscript{\textasteriskcentered}}         & .5624 & .5136 & .5790 & .6358 & .5304 & .5972 & .5705 & .5882 & .5642 & .5912 \\ \hline
        \multicolumn{1}{l|}{ContraCLIP} & \textbf{.8662} & \textbf{.9245} & \textbf{.9099} & \textbf{.9559} & \textbf{.9372} & \textbf{.9407} & \textbf{.9402} & \textbf{.9426} & \textbf{.9344} & \textbf{.9567} \\ \hline
        \end{tabular}
        \label{tab:id_results}
    \end{table}

\section{GAN pre-trained on non-facial datasets}\label{supp:other_datasets}
\begin{figure}[t!]
        \centering
        \includegraphics[width=0.95\textwidth]{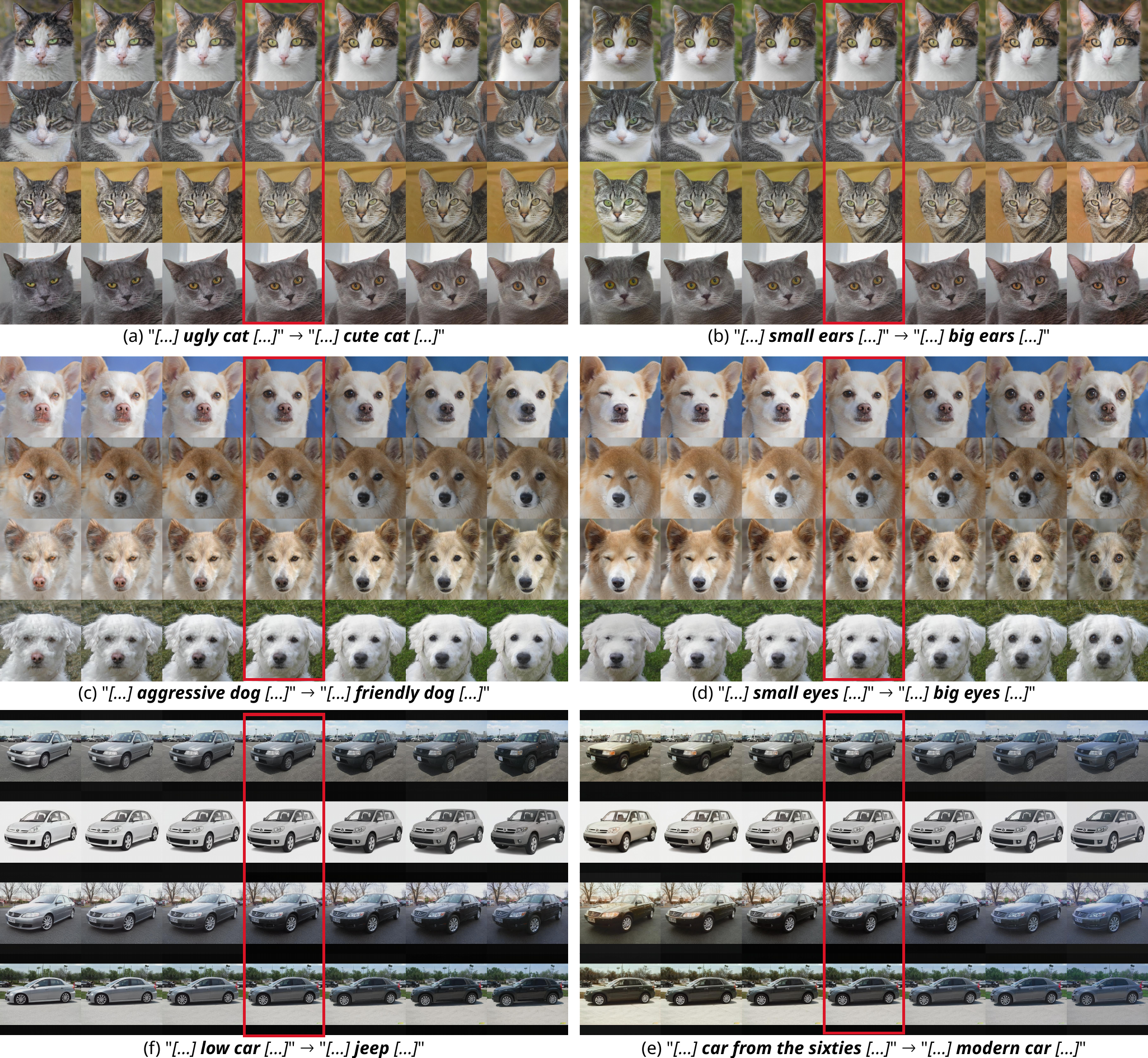}
        \caption{Results of the proposed method on non-facial GANs, i.e., StyleGAN2 pretrained on (a,b) AFHQ Cats~\cite{choi2020stargan}, (c,d) AFHQ Dogs~\cite{choi2020stargan}, and (e,f) LSUN Cars~\cite{yu2015lsun}.}
        \label{fig:cats_dogs_cars}
    \end{figure}

    Besides the fact that the proposed method is model-agnostic and not tied to any specific GAN architecture (e.g., StyleGAN), such as~\cite{patashnik2021styleclip,xia2021tedigan}, it is also not limited to any specific type of imagery (i.e., facial images), such as~\cite{xia2021tedigan}. To support the former claim, we conducted and reported in Sect.~\ref{sec:experiments} experiments using different GAN architectures (i.e., ProgGAN~\cite{proggan_karras18iclr} and StyleGAN2~\cite{stylegan2_karras20cvpr}). To support the latter claim, we conducted and report in this section additional experiments using StyleGAN2 pre-trained on AFHQ Cats~\cite{choi2020stargan}, AFHQ Dogs~\cite{choi2020stargan}, and LSUN Cars~\cite{yu2015lsun}. We show the results in Fig.~\ref{fig:cats_dogs_cars}. We observe that the proposed method, i.e., the learnt non-linear latent paths driven by non-linear paths in the CLIP text space, arrive at high quality images and manipulations that are largely consistent with the semantics expressed in the natural language sentences that we defined.
    
\end{appendices}

\bibliographystyle{splncs04}
\bibliography{preprint}

\begin{thebibliography}{10}
\providecommand{\url}[1]{\texttt{#1}}
\providecommand{\urlprefix}{URL }
\providecommand{\doi}[1]{https://doi.org/#1}

\bibitem{abdal2021styleflow}
Abdal, R., Zhu, P., Mitra, N.J., Wonka, P.: Styleflow: Attribute-conditioned
  exploration of stylegan-generated images using conditional continuous
  normalizing flows. ACM Transactions on Graphics (TOG)  \textbf{40}(3),  1--21
  (2021)

\bibitem{brown2020language}
Brown, T., Mann, B., Ryder, N., Subbiah, M., Kaplan, J.D., Dhariwal, P.,
  Neelakantan, A., Shyam, P., Sastry, G., Askell, A., et~al.: Language models
  are few-shot learners. Advances in neural information processing systems
  \textbf{33},  1877--1901 (2020)

\bibitem{chen2020uniter}
Chen, Y.C., Li, L., Yu, L., El~Kholy, A., Ahmed, F., Gan, Z., Cheng, Y., Liu,
  J.: Uniter: Universal image-text representation learning. In: European
  conference on computer vision. pp. 104--120. Springer (2020)

\bibitem{choi2020stargan}
Choi, Y., Uh, Y., Yoo, J., Ha, J.W.: Stargan v2: Diverse image synthesis for
  multiple domains. In: Proceedings of the IEEE/CVF conference on computer
  vision and pattern recognition. pp. 8188--8197 (2020)

\bibitem{deng2019arcface}
Deng, J., Guo, J., Xue, N., Zafeiriou, S.: Arcface: Additive angular margin
  loss for deep face recognition. In: Proceedings of the IEEE/CVF Conference on
  Computer Vision and Pattern Recognition. pp. 4690--4699 (2019)

\bibitem{desai2021virtex}
Desai, K., Johnson, J.: Virtex: Learning visual representations from textual
  annotations. In: Proceedings of the IEEE/CVF Conference on Computer Vision
  and Pattern Recognition. pp. 11162--11173 (2021)

\bibitem{devlin2018bert}
Devlin, J., Chang, M.W., Lee, K., Toutanova, K.: {BERT}: Pre-training of deep
  bidirectional transformers for language understanding. arXiv preprint
  arXiv:1810.04805  (2018)

\bibitem{dosovitskiy2020image}
Dosovitskiy, A., Beyer, L., Kolesnikov, A., Weissenborn, D., Zhai, X.,
  Unterthiner, T., Dehghani, M., Minderer, M., Heigold, G., Gelly, S., et~al.:
  An image is worth 16x16 words: Transformers for image recognition at scale.
  arXiv preprint arXiv:2010.11929  (2020)

\bibitem{goetschalckx2019ganalyze}
Goetschalckx, L., Andonian, A., Oliva, A., Isola, P.: Ganalyze: Toward visual
  definitions of cognitive image properties. In: Proceedings of the IEEE/CVF
  International Conference on Computer Vision. pp. 5744--5753 (2019)

\bibitem{goodfellow2014generative}
Goodfellow, I., Pouget-Abadie, J., Mirza, M., Xu, B., Warde-Farley, D., Ozair,
  S., Courville, A., Bengio, Y.: Generative adversarial nets. Advances in
  neural information processing systems  \textbf{27} (2014)

\bibitem{ganspace2020harkonen}
H{\"a}rk{\"o}nen, E., Hertzmann, A., Lehtinen, J., Paris, S.: Ganspace:
  Discovering interpretable gan controls. Advances in Neural Information
  Processing Systems  \textbf{33},  9841--9850 (2020)

\bibitem{jahanian20iclr}
Jahanian, A., Chai, L., Isola, P.: On the "steerability" of generative
  adversarial networks. In: 8th International Conference on Learning
  Representations, {ICLR} 2020, Addis Ababa, Ethiopia, April 26-30, 2020.
  OpenReview.net (2020)

\bibitem{talktoedit2021iccv}
Jiang, Y., Huang, Z., Pan, X., Loy, C.C., Liu, Z.: Talk-to-edit: Fine-grained
  facial editing via dialog. In: Proceedings of the IEEE/CVF International
  Conference on Computer Vision. pp. 13799--13808 (2021)

\bibitem{proggan_karras18iclr}
Karras, T., Aila, T., Laine, S., Lehtinen, J.: Progressive growing of gans for
  improved quality, stability, and variation. In: 6th International Conference
  on Learning Representations, {ICLR} 2018, Vancouver, BC, Canada, April 30 -
  May 3, 2018, Conference Track Proceedings (2018)

\bibitem{stylegan2_karras20cvpr}
Karras, T., Laine, S., Aittala, M., Hellsten, J., Lehtinen, J., Aila, T.:
  Analyzing and improving the image quality of stylegan. In: 2020 {IEEE/CVF}
  Conference on Computer Vision and Pattern Recognition, {CVPR} 2020, Seattle,
  WA, USA, June 13-19, 2020. pp. 8107--8116. {IEEE} (2020)

\bibitem{ledig2017photo}
Ledig, C., Theis, L., Husz{\'a}r, F., Caballero, J., Cunningham, A., Acosta,
  A., Aitken, A., Tejani, A., Totz, J., Wang, Z., et~al.: Photo-realistic
  single image super-resolution using a generative adversarial network. In:
  Proceedings of the IEEE conference on computer vision and pattern
  recognition. pp. 4681--4690 (2017)

\bibitem{li2020manigan}
Li, B., Qi, X., Lukasiewicz, T., Torr, P.H.: Manigan: Text-guided image
  manipulation. In: Proceedings of the IEEE/CVF Conference on Computer Vision
  and Pattern Recognition. pp. 7880--7889 (2020)

\bibitem{li2020unicoder}
Li, G., Duan, N., Fang, Y., Gong, M., Jiang, D.: Unicoder-vl: A universal
  encoder for vision and language by cross-modal pre-training. In: Proceedings
  of the AAAI Conference on Artificial Intelligence. vol.~34, pp. 11336--11344
  (2020)

\bibitem{li2020oscar}
Li, X., Yin, X., Li, C., Zhang, P., Hu, X., Zhang, L., Wang, L., Hu, H., Dong,
  L., Wei, F., et~al.: Oscar: Object-semantics aligned pre-training for
  vision-language tasks. In: European Conference on Computer Vision. pp.
  121--137. Springer (2020)

\bibitem{celeba_liu15iccv}
Liu, Z., Luo, P., Wang, X., Tang, X.: Deep learning face attributes in the
  wild. In: 2015 {IEEE} International Conference on Computer Vision, {ICCV}
  2015, Santiago, Chile, December 7-13, 2015. pp. 3730--3738. {IEEE} Computer
  Society (2015)

\bibitem{lu2019vilbert}
Lu, J., Batra, D., Parikh, D., Lee, S.: Vilbert: Pretraining task-agnostic
  visiolinguistic representations for vision-and-language tasks. Advances in
  neural information processing systems  \textbf{32} (2019)

\bibitem{oldfield2021tensor}
Oldfield, J., Georgopoulos, M., Panagakis, Y., Nicolaou, M.A., Patras, I.:
  Tensor component analysis for interpreting the latent space of gans. arXiv
  preprint arXiv:2111.11736  (2021)

\bibitem{patashnik2021styleclip}
Patashnik, O., Wu, Z., Shechtman, E., Cohen-Or, D., Lischinski, D.:
  {StyleCLIP}: Text-driven manipulation of stylegan imagery. In: Proceedings of
  the IEEE/CVF International Conference on Computer Vision. pp. 2085--2094
  (2021)

\bibitem{plumerault20iclr}
Plumerault, A., Borgne, H.L., Hudelot, C.: Controlling generative models with
  continuous factors of variations. In: 8th International Conference on
  Learning Representations, {ICLR} 2020, Addis Ababa, Ethiopia, April 26-30,
  2020. OpenReview.net (2020), \url{https://openreview.net/forum?id=H1laeJrKDB}

\bibitem{clip-radford2021icml}
Radford, A., Kim, J.W., Hallacy, C., Ramesh, A., Goh, G., Agarwal, S., Sastry,
  G., Askell, A., Mishkin, P., Clark, J., et~al.: Learning transferable visual
  models from natural language supervision. In: International Conference on
  Machine Learning. pp. 8748--8763. PMLR (2021)

\bibitem{ramesh2021zero}
Ramesh, A., Pavlov, M., Goh, G., Gray, S., Voss, C., Radford, A., Chen, M.,
  Sutskever, I.: Zero-shot text-to-image generation. In: International
  Conference on Machine Learning. pp. 8821--8831. PMLR (2021)

\bibitem{reed2016generative}
Reed, S., Akata, Z., Yan, X., Logeswaran, L., Schiele, B., Lee, H.: Generative
  adversarial text to image synthesis. In: International conference on machine
  learning. pp. 1060--1069. PMLR (2016)

\bibitem{sariyildiz2020learning}
Sariyildiz, M.B., Perez, J., Larlus, D.: Learning visual representations with
  caption annotations. In: European Conference on Computer Vision. pp.
  153--170. Springer (2020)

\bibitem{shen2020interpreting}
Shen, Y., Gu, J., Tang, X., Zhou, B.: Interpreting the latent space of gans for
  semantic face editing. In: Proceedings of the IEEE/CVF Conference on Computer
  Vision and Pattern Recognition. pp. 9243--9252 (2020)

\bibitem{shen2020interfacegan}
Shen, Y., Yang, C., Tang, X., Zhou, B.: Interfacegan: Interpreting the
  disentangled face representation learned by gans. IEEE transactions on
  pattern analysis and machine intelligence  (2020)

\bibitem{sefa2021cvpr}
Shen, Y., Zhou, B.: Closed-form factorization of latent semantics in gans. In:
  Proceedings of the IEEE/CVF Conference on Computer Vision and Pattern
  Recognition. pp. 1532--1540 (2021)

\bibitem{warpedganspace2021iccv}
Tzelepis, C., Tzimiropoulos, G., Patras, I.: {WarpedGANSpace}: Finding
  non-linear rbf paths in {GAN} latent space. In: Proceedings of the IEEE/CVF
  International Conference on Computer Vision (ICCV). pp. 6393--6402 (October
  2021)

\bibitem{vaswani2017attention}
Vaswani, A., Shazeer, N., Parmar, N., Uszkoreit, J., Jones, L., Gomez, A.N.,
  Kaiser, {\L}., Polosukhin, I.: Attention is all you need. Advances in neural
  information processing systems  \textbf{30} (2017)

\bibitem{voynov2020unsupervised}
Voynov, A., Babenko, A.: Unsupervised discovery of interpretable directions in
  the gan latent space. In: International conference on machine learning. pp.
  9786--9796. PMLR (2020)

\bibitem{wu2021stylespace}
Wu, Z., Lischinski, D., Shechtman, E.: Stylespace analysis: Disentangled
  controls for stylegan image generation. In: Proceedings of the IEEE/CVF
  Conference on Computer Vision and Pattern Recognition. pp. 12863--12872
  (2021)

\bibitem{xia2021tedigan}
Xia, W., Yang, Y., Xue, J.H., Wu, B.: {TediGAN}: Text-guided diverse face image
  generation and manipulation. In: Proceedings of the IEEE/CVF Conference on
  Computer Vision and Pattern Recognition. pp. 2256--2265 (2021)

\bibitem{xu2018attngan}
Xu, T., Zhang, P., Huang, Q., Zhang, H., Gan, Z., Huang, X., He, X.: {AttnGAN}:
  Fine-grained text to image generation with attentional generative adversarial
  networks. In: Proceedings of the IEEE conference on computer vision and
  pattern recognition. pp. 1316--1324 (2018)

\bibitem{xu2021generative}
Xu, Y., Shen, Y., Zhu, J., Yang, C., Zhou, B.: Generative hierarchical features
  from synthesizing images. In: Proceedings of the IEEE/CVF Conference on
  Computer Vision and Pattern Recognition. pp. 4432--4442 (2021)

\bibitem{yang2021gan}
Yang, T., Ren, P., Xie, X., Zhang, L.: Gan prior embedded network for blind
  face restoration in the wild. In: Proceedings of the IEEE/CVF Conference on
  Computer Vision and Pattern Recognition. pp. 672--681 (2021)

\bibitem{yu2015lsun}
Yu, F., Seff, A., Zhang, Y., Song, S., Funkhouser, T., Xiao, J.: Lsun:
  Construction of a large-scale image dataset using deep learning with humans
  in the loop. arXiv preprint arXiv:1506.03365  (2015)

\end{thebibliography}

\end{document}